# The Myth of Modularity in Rule-Based Systems

David E. Heckerman and Eric J. Horvitz

Medical Computer Science Group
Knowledge Systems Laboratory
Stanford University
Medical School Office Building, Room 215
Stanford, California 94305

## Abstract

In this paper, we examine the concept of *modularity*, an often cited advantage of the ruled-based representation methodology. We argue that the notion of modularity consists of two distinct concepts which we call *syntactic* modularity and *semantic* modularity. We argue that when reasoning under uncertainty, it is reasonable to regard the rule-based approach as both syntactically and semantically modular. However, we argue that in the case of plausible reasoning, rules are syntactically modular but are rarely semantically modular. To illustrate this point, we examine a particular approach for managing uncertainty in rule-based systems called the MYCIN certainty factor model. We formally define the concept of semantic modularity with respect to the certainty factor model and discuss logical consequences of the definition. We show that the assumption of semantic modularity imposes strong restrictions on rules in a knowledge base; we argue that such restrictions are rarely valid in practical applications. Finally, we suggest how the concept of semantic modularity can be relaxed in a manner that makes it appropriate for plausible reasoning.

## Introduction

Researchers in the artificial intelligence community have concentrated their efforts on deductive problem solving methods. In doing so, they have developed numerous approaches for representing and manipulating propositions. One methodology that has been used frequently to build expert systems is the *rule-based* representation framework. In rule-based systems, knowledge is represented as rules of the form "IF A THEN B" where A and B are logical propositions.

An often cited advantage of the rule-based approach is that rules can be added or deleted from a knowledge base without the need to modify other rules [4]. This property is called *modularity*. To our knowledge the concept of modularity has never been formally defined. Nevertheless, modularity has been informally described in some detail. For example, the following two paragraphs are taken from a discussion of modularity by Davis [4]:

> We can regard the *modularity* of a program as the degree of separation of its functional units into isolatable pieces. A program is *highly modular* if any functional unit can be changed (added, deleted, or replaced) with no unanticipated change to other functional units. Thus program modularity is inversely related to the strength of coupling between its functional units.
>
> The modularity of programs written as pure production systems arises from the important fact that the next rule to be invoked is determined solely by the contents of the data base, and no rule is ever called directly. Thus the addition (or deletion) of a rule does not require the modification of any other rule to provide for or delete a call to it. We might demonstrate this by repeatedly removing rules from a PS [production system]: many systems will continue to display some sort of "reasonable" behavior. By contrast, adding a procedure to an ALGOL-like program requires modification of other parts of the code to insure that the procedure is invoked, while removing an arbitrary procedure from such a program will generally cripple it.

In the above quotation and in other discussions of modularity, it seems that two different notions of modularity are defined without apparent distinction. One notion is that rules can be added or deleted from a knowledge base without altering the *truth* or *validity* of other rules in the system. The other notion is that rules can be added or deleted from a knowledge base without modifying the *syntax* of other rules; the inference process can continue in spite of such additions or deletions. We will call the former notion *semantic modularity* and the latter *syntactic modularity*.

By design, rules are syntactically modular. Furthermore, the validity of a rule in a deductive system does not depend on other rules in the system's knowledge base. For example, if a knowledge base contains the rule

```
IF:   A and B are parallel lines
THEN: A and B do not intersect
```

then the addition or deletion of other rules will not affect the truth value of this rule. Once this fact is asserted, it cannot be falsified by additional facts. Of course, a rule might be added that contradicts rules already in the knowledge base. Aside from this special case, however, categorical rules are semantically modular. Indeed, the syntactically modular rule-based representation scheme may have emerged from the recognition that rules are modular in the semantic sense.

As investigators have begun to tackle real-world problems such as mineral exploration, medical diagnosis, and financial management, methods for reasoning under uncertainty or *plausible reasoning* have received increasing attention. Popular AI approaches that have been developed for managing uncertainty include extensions of the production rule methodology. In these methodologies, a number is attached to each rule which represents the degree of association, in some sense, between the antecedent and the consequent of the rule.

In such approaches, the notions of syntactic and semantic modularity have been carried over from deductive systems. That is, the properties of both syntactic and semantic modularity have been ascribed to rules in plausible reasoning systems. It seems appropriate to attribute the property of syntactic modularity to rules in plausible reasoning systems. Just as in deductive systems, non-categorical rules can be added and deleted without the need to modify the syntax of other rules. However, in this paper, we argue that it is inappropriate to carry over the notion of semantic modularity from deductive systems and apply it to systems which *must* manage uncertainty. We shall see that fundamental differences between logical and plausible reasoning result in the breakdown of the assumption of semantic modularity in rule-based systems which reason under uncertainty.

To demonstrate that it is inappropriate to ascribe the property of *semantic* modularity to rules in plausible reasoning systems, we will examine a particular rule-based method for reasoning under uncertainty, the MYCIN certainty factor (CF) model [5]. We will first present a formal definition of semantic modularity with respect to the CF model. We will then illustrate several implications of semantic modularity and argue that these implications cannot be justified in most practical applications. Also in this paper, we will discuss a methodology for relaxing the assumption of semantic modularity to accommodate plausible reasoning.

We should emphasize that we are not the first to notice problems with the assumption of modularity in rule-based systems which reason under uncertainty. For example, in the closing remarks of their book on rule-based expert systems [6], Buchanan and Shortliffe state that many of MYCIN's rules do not have the property that we have termed semantic modularity. However to our knowledge, there have been no efforts to formally define the concept of semantic modularity nor have there been efforts to contrast the two concepts of syntactic and semantic modularity in detail.

## Overview of the MYCIN certainty factor model

In this section, we will describe the aspects of the MYCIN CF model that are essential for understanding the consequences of modularity in rule-based systems which manage uncertainty. As mentioned above, the model is an adjunct to the rule-based representation framework. A *certainty factor* is attached to each rule that represents the *change in belief* in consequent of the rule given the antecedent. Certainty factors range between -1 and 1. Positive numbers correspond to an *increase* in belief in a hypothesis while negative quantities correspond to a *decrease* in belief.[1] It is important to note that certainty factors do not correspond to measures of *absolute* belief. This distinction, with respect to certainty factors as well as other measures of uncertainty, has not been emphasized in the artificial intelligence literature [7].

The following notation will be used to represent a rule in the CF model:

$$E \xrightarrow{CF(H,E)} H$$

where H is a hypothesis, E is evidence relating to the hypothesis, and CF(H,E) is the certainty factor attached to the rule. In MYCIN, multiple pieces of evidence may bear on the same hypothesis. In addition, a hypothesis may serve as evidence for yet another hypothesis. The result is a network of rules such as the one shown in Figure 1. This structure is often called an *inference network* [8]. The certainty factor model includes a prescription for propagating uncertainty through an inference network. For example, the CF model can be used to compute the change in belief in hypotheses G and H when A and B are true (see Figure 1). Details of the propagation scheme are described in [5] and [6].

## Definition of semantic modularity

In this section, we construct a formal definition of semantic modularity in the context of the MYCIN certainty factor model. To motivate the definition, consider the following classic example from probability theory. Suppose an individual is given one of the following two urns:

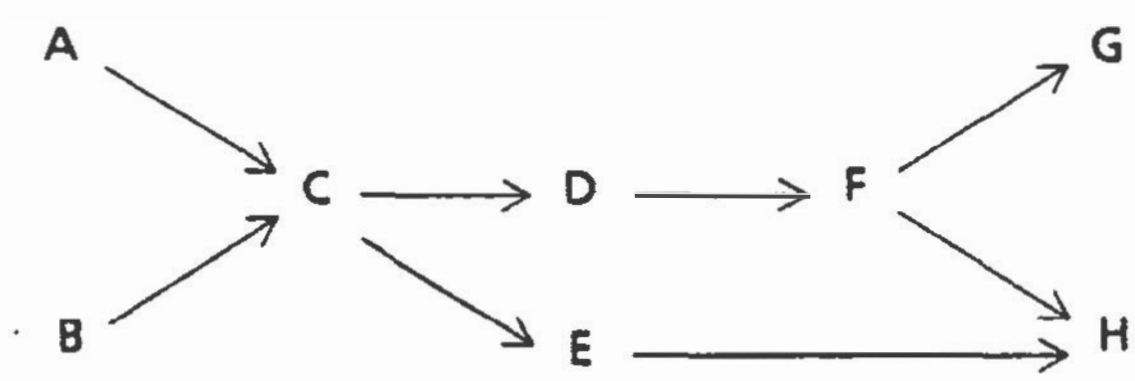

**Figure 1:** An inference network

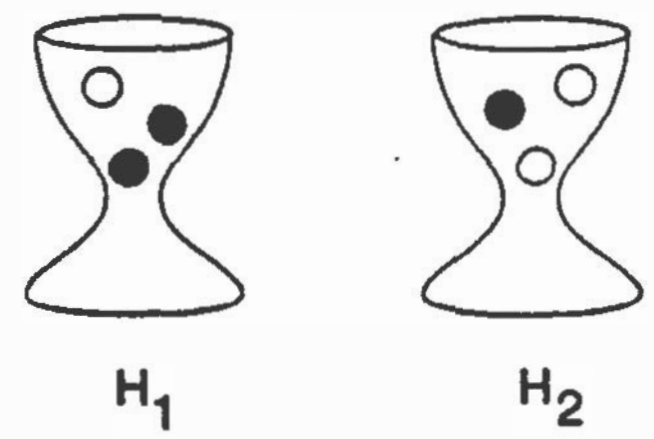

The first urn contains 1 white ball and 2 black balls while the second urn contains 2 white balls and 1 black ball. He is not allowed to look inside the urn, but is allowed to draw balls from the urn, one at a time, *without* replacement. Let $H_1$ be the hypothesis that the individual is holding the first urn and $H_2$ be the hypothesis that he is holding the second urn.

Suppose a black ball and then a second black ball are drawn from the urn. Upon drawing the second black ball, the individual's belief in $H_1$ increases to certainty. In contrast, suppose the result of the first draw is a white ball. In this case, the draw of the second black ball raises his belief in $H_1$ somewhat but does not confirm the hypothesis. Therefore, the effect of the second draw on his belief about the identity of the urn is strongly dependent on the result of the first draw.

This example illustrates that changes in belief may depend on information available at the time evidence becomes known. Therefore, a certainty factor which represents the change in belief in a hypothesis H given evidence E should be written as a function of three arguments, CF(H,E,e), where e denotes information that is known at the time of the update. The notion of semantic modularity, however, requires that the certainty factor for the rule "IF E THEN H" *not* depend on whether other rules which bear on H have fired. Therefore, we take as the formal definition of semantic modularity:

$$CF(H,E,e) = CF(H,E,\emptyset) \qquad (1)$$

for all H and E in the network and for any evidence e that might be known at the time E updates H.[2] Since certainty factors in the CF model are functions of only two arguments, the model implicitly assumes (1).

## Consequences of modularity

Although little can be deduced from the modularity property alone, the creators of the CF model, in order to justify its use, outlined a set of properties or *desiderata* that should be satisfied by a propagation scheme. For example, one desideratum requires that the order in which evidence is considered should not affect the result of propagation. These desiderata provide tools that can be used to deduce the consequences of modularity.

In fact, we note that the desiderata alone lead to a significant consequence. It can be shown that *any* quantity which satisfies the desiderata *must* be a monotonic transformation of the the likelihood ratio $\lambda(H,E,e) = p(E|H\wedge e)/p(E|\sim H\wedge e)$, where $p(E|H\wedge e)$ is the *probability* of evidence E given hypothesis H in the context of information e and $p(E|\sim H\wedge e)$ is the probability of evidence E given hypothesis H is false. That is, $CF(H,E,e) = F(\lambda(H,E,e))$ for some monotonic increasing function F. For example, the mapping

$$CF(H,E,e) = \begin{cases} (\lambda(H,E,e)-1)/\lambda(H,E,e) & \lambda \geq 1 \\ \lambda(H,E,e)-1 & \lambda < 1 \end{cases} \qquad (2)$$

can be shown to be consistent with one of the central methods of propagation used in the CF model [9].

We will not present a formal proof of the above nor proofs of the consequences of semantic modularity in this paper. Instead, we will state several consequences of modularity and provide examples that attempt to convey an intuitive understanding of the proofs. Those interested in the proofs should see [9].

__Conditional independence__

The first consequence we discuss concerns a common situation where several pieces of evidence bear on a single hypothesis. This is shown below:

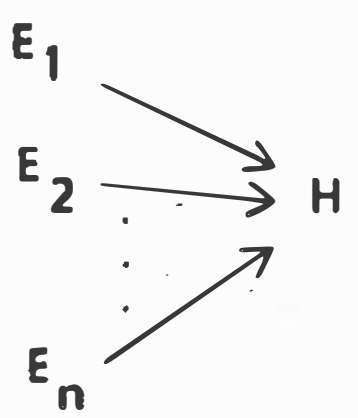

Let $\mathcal{E}_H$ denote the set $\{E_1, E_2, \ldots E_n\}$.[3] Consider a single item of evidence $E_i$ in $\mathcal{E}_H$ and let e be any subset of $\mathcal{E}_H$ which does not include $E_i$. In this situation, it can be proved that the modularity property (1) holds if and only if

$$p(E_i|H\wedge e) = p(E_i|H) \quad \text{and} \qquad (3)$$

$$p(E_i|\sim H\wedge e) = p(E_i|\sim H).$$

Equation (3) says that the belief in $E_i$ does not depend on the knowledge that e is true when H is definitely true or definitely false. When (3) holds, it is said that evidence is *conditionally independent* given H and its negation.

Let us consider this correspondence between conditional independence and the modularity property in the context of the urn problem above. In the example, it is a simple matter to see why the modularity property is violated; draws are done *without* replacement, making evidence conditionally *dependent*. For example,

$$p(\text{2nd draw black}|H_2\wedge\text{1st draw black}) = 0$$

and

$$p(\text{2nd draw black}|H_2\wedge\text{1st draw white}) = 1/2.$$

Clearly, condition (3) is not satisfied in the example; this is consistent with our previous observation that modularity does not hold.

The urn problem can be modified such that the modularity property is satisfied. If draws are done *with* replacement, the conditional independence condition (3) is satisfied. In particular, for any e:

$$p(W|H_1\wedge e) = p(W|H_1) = 1/3$$

$$p(W|\sim H_1\wedge e) = p(W|H_2\wedge e) = p(W|H_2) = 2/3$$

where W denotes the draw of a white ball and e denotes draws made prior to W. Since (3) is satisfied, the modularity property holds and we can compute $CF(H_1,W)$, a function of only two arguments, using relation (2):

$$\lambda(H_1,W) = (1/3)/(2/3) = .5$$

$$CF(H_1,W) = .5 - 1 = -.5.$$

Other certainty factors relevant to the problem can be calculated in a similar fashion.

Unfortunately, semantic modularity only holds in extremely simple situations like the one above. Any small increase in the complexity of the problem will result in the loss of modularity. For example, suppose an individual is given one of *three* urns:

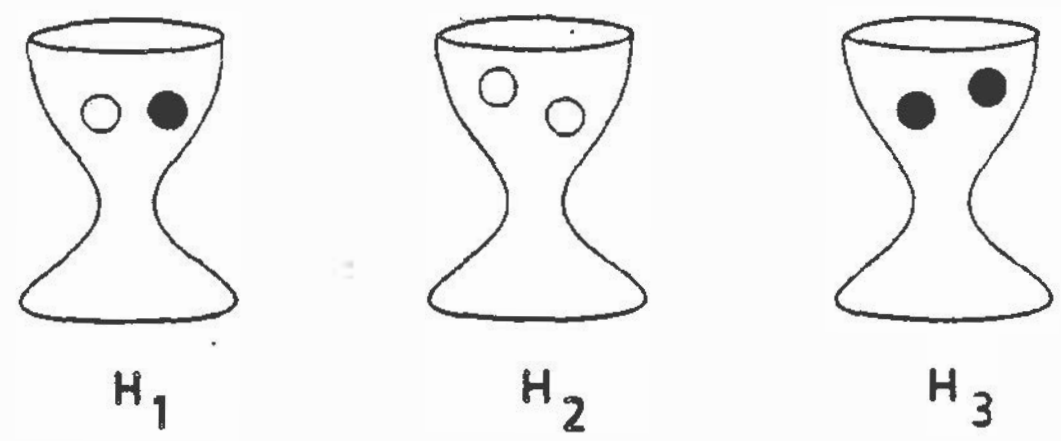

Making draws *with* replacement, evidence is conditionally independent given each of the hypotheses $H_1$, $H_2$, and $H_3$. However, evidence is no longer conditionally independent given the *negation* of any hypothesis. For example, if each hypothesis is equally likely before any draws, the initial probability of drawing a black ball given $\sim H_1$ is

$$\begin{aligned} p(B|\sim H_1) &= p(B\wedge\sim H_1) \;/\; p(\sim H_1) \\ &= [p(B\wedge H_2) + p(B\wedge H_3)] \;/\; p(\sim H_1) \\ &= [p(B|H_2)p(H_2) + p(B|H_3)p(H_3)] \;/\; p(\sim H_1) \\ &= [(0)(1/3) + (1)(1/3)] \;/\; (2/3) \\ &= 1/2 \end{aligned}$$

However, if a white ball is drawn, $H_3$ is ruled out and the probability of drawing a black ball changes to

$$p(B|\sim H_1\wedge W) = p(B|H_2) = 0.$$

Given the correspondence between the conditional independence assumption (3) and the modularity property (1) cited above, it follows that the rules describing this situation cannot be semantically modular. Indeed, using (2) we find that

$$\lambda(H_1,B,\emptyset) = 1 \quad ==> \quad CF(H_1,B,\emptyset) = 0$$

$$\lambda(H_1,B,W) = \infty \quad ==> \quad CF(H_1,B,W) = 1$$

and therefore

$$CF(H_1,B,\emptyset) \neq CF(H_1,B,W).$$

Intuitively, if a black ball is drawn first, one's belief in $H_1$ does not change significantly. However, if a black ball is drawn following the draw of a white ball, $H_3$ is ruled out and $H_1$ is confirmed. Thus, the certainty factor for $H_1$ depends on other pieces of evidence (other rules in the inference network). Consequently, the rules representing this knowledge are not modular.

The lack of modularity can be traced directly to the fact that there are more than two mutually exclusive and exhaustive events. In such cases, $\sim H_i$ is a "mixture" of hypotheses and thus evidence will not be conditionally independent given $\sim H_i$ even when evidence is conditionally independent given each $H_i$. Since the conditional independence assumption (3) is not satisfied, the modularity property cannot hold. This result can be rigorously derived. It can be shown that whenever a set of mutually exclusive and exhaustive hypotheses contains more than two elements, the conditional independence assumption (3) is incompatible with multiple updating [10].

We must emphasize the severe nature of this implication of the conditional independence assumption. Clearly, it is seldom true in real-world applications. Furthermore, given the correspondence between (3) and (1), it follows that the assumption of semantic modularity is at least as restrictive.

### A restriction on the topology of inference networks

Another restrictive consequence of semantic modularity is that evidence *cannot*, in most circumstances, be propagated in a consistent manner through networks in which a single piece of evidence bears on more than one hypothesis[4] as shown below:

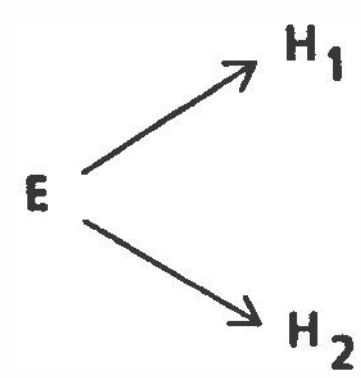

Notice the asymmetry between the case of *convergent* links discussed above and this case of *divergent* links. In the former case, the propagation of uncertainty is possible provided conditional independence is assumed. In the latter case, propagation in a manner consistent with the modularity property is essentially impossible.

To illustrate the difficulty associated with divergent links, consider the following story due to Kim and Pearl [11]:

> Mr. Holmes received a telephone call from his neighbor notifying him that she heard a burglar alarm sound from the direction of his home. As he was preparing to rush home, Mr. Holmes recalled that last time the alarm had been triggered by an earthquake. On his way driving home, he heard a radio newscast reporting an earthquake 200 miles away.

An inference network corresponding to this story is shown in Figure 2.

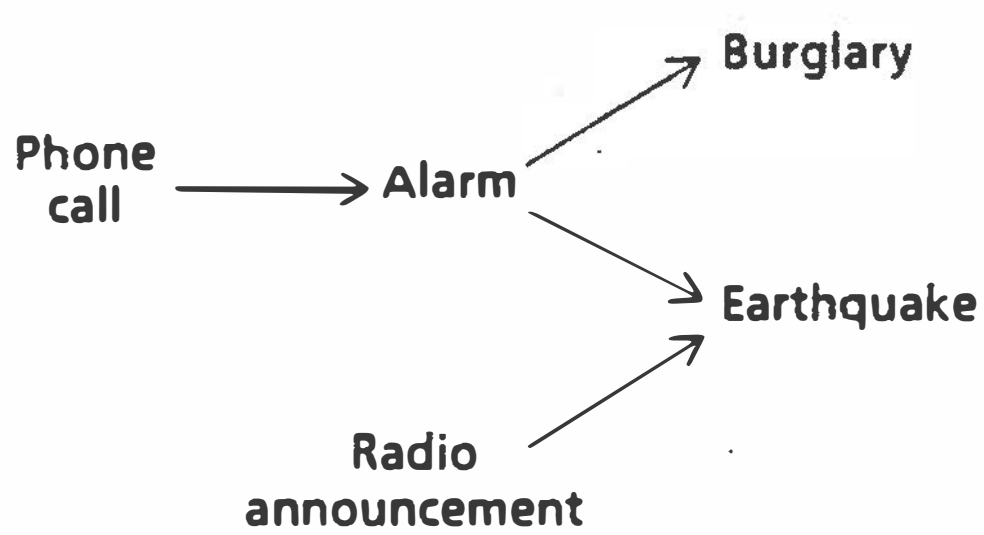

**Figure 2:** An inference network for Mr. Holmes' situation

A problem arises in trying to assign certainty factors to the rules which have "Alarm" as the antecedent. Since Mr. Holmes heard the radio announcement, the alarm sound tends to support the *earthquake* hypothesis. However, had Mr. Holmes not heard the radio announcement, the alarm sound would lend more support to the *burglary* hypothesis. Thus, the rules above are not modular since their impact depends on the belief about another proposition in the network.

There is an intuitive explanation for this lack of modularity. *A priori*, the earthquake hypothesis and burglary hypothesis are independent. However, knowledge of the alarm status induces a dependency between these hypotheses. In particular, once Mr. Holmes knows the alarm has sounded, evidence *for* one hypothesis is indirect evidence *against* the other. This dependency couples the rules in such a way that modularity is lost.

In this scenario, the loss of modularity is consistent with the general result cited above. Because the network contains divergent links ("IF Alarm THEN Burglary" and "IF Alarm THEN Earthquake"), updating is only possible if the semantic modularity property, (1), does not hold.

## The myth of modularity

As noted above, extremely few expert system domains satisfy the severely restrictive consequences of semantic modularity described in the previous section. Indeed, it is difficult to imagine a domain in which all sets of mutually exclusive and exhaustive hypotheses contain only two elements and in which divergent links do not occur. It follows, therefore, that semantic modularity rarely exists in rule-based systems which use the CF model to manage uncertainty. Earlier, however, we argued that rules in deductive systems (CF's = $\pm 1$) are semantically modular.

The distinction between categorical and non-categorical rules is suggested by a fundamental difference between deductive and plausible reasoning alluded to earlier. Suppose a hypothesis is believed with certainty. In this case, no additional information can refute its truth. In particular, if a knowledge base contains the rule "IF E THEN H" (CF = 1), and if E is categorically established, then H is also established beyond any doubt. The addition of other rules to the knowledge base or the establishment of other rule antecedents cannot affect this conclusion. However, if a hypothesis is uncertain, the degree to which it is believed is sensitive to new information. If a knowledge base contains the rule "IF E THEN H" (CF $\neq \pm 1$), then additional information such as new evidence for H can affect the certainty factor for this rule.

In general, the strength of association between antecedent and consequent in a non-categorical rule will change when other rules are added to or deleted from a knowledge base. This suggests that semantic modularity (in a more general sense) will rarely hold in any rule-based system which must manage uncertainty. Thus, it appears that non-categorical rules will maintain syntactic modularity but *not* semantic modularity. The notion that syntactic and semantic modularity go hand in hand is a myth.

## A weaker notion of modularity

Since the assumption of semantic modularity places severe constraints on the relationships among propositions in a knowledge base, it seems useful to modify the ruled-based representation scheme to accommodate a weaker form of semantic modularity that is appropriate for plausible reasoning. In this section, we examine a methodology for representing and propagating uncertainty called *influence diagrams* [12] and show that this approach suggests a weaker form of modularity suited to plausible reasoning.[5]

We first informally describe the influence diagram methodology. In doing so, we show how this approach is used to represent the example problems discussed above. We next contrast the influence diagram methodology with the inference network approach. Finally, we develop the weaker notion of semantic modularity suggested by the approach.

An influence diagram is a two-level structure. The upper level of an influence diagram consists of a graph that represents the propositions relevant to a problem as well as relationships among them. Nodes (circles) are used to represent propositions and directed arcs are used to represent dependencies among the propositions. The bottom level represents all possible values or outcomes for each proposition together with a *probability distribution* for each proposition. The arcs in the upper level represent the notion of probabilistic conditioning or *influence*. In particular, an arc from proposition A to proposition B means that the probability distribution for B can be influenced by the values of A. If there is no arc from A to B, the probability distribution for B is independent of the values of A. Thus, the influence diagram representation is useful for representing assumptions of conditional independence. Notice that an arc from A to B may be interpreted as either a *causal* influence or an *associational* influence; no distinction is made between these two concepts in an influence diagram.

To illustrate these concepts, consider once again the three urn problem. An influence diagram for this problem is shown in Figure 3.

**Level 1:**

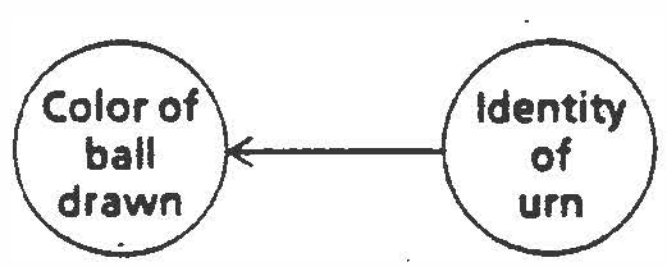

**Level 2:**

| Node: Color of ball drawn | | |
|---|---|---|
| Identity of urn | Values | p(Color \| Identity) |
| $H_1$ | White | 1/2 |
| | Black | 1/2 |
| $H_2$ | White | 1 |
| | Black | 0 |
| $H_3$ | White | 0 |
| | Black | 1 |

| Node: Identity of urn | |
|---|---|
| Values | p(Identity) |
| $H_1$ | 1/3 |
| $H_2$ | 1/3 |
| $H_3$ | 1/3 |

**Figure 3:** An influence diagram for the three urn problem

The two nodes labeled "Identity of urn" and "Color of ball drawn" in the upper level of the diagram represent the propositions relevant to the problem. The tables in the lower level list the possible values for each proposition. The arc between the two nodes in the upper level means that the probability distribution for "Color of ball drawn" depends on the value of "Identity of urn." Consequently, the probability distribution for "Color of ball drawn" given in the second level of the diagram is conditioned on each of the three possible values of "Identity of urn": $H_1$, $H_2$, and $H_3$. Note that since there are no arcs into the "Identity of urn" node an unconditional or *marginal* distribution for this proposition is given. Also note that the same urn problem can be represented by an influence diagram with the arc reversed. In this case, a marginal probability distribution would be assigned to "Color of ball drawn" and a conditional probability distribution would be assigned to "Identity of urn."

We see that there are several significant differences between influence diagrams and inference networks. The first difference is that an influence diagram is a two-level structure while the inference network contains only one level. Another difference is that propositions in an influence diagram can take on any number (possibly infinite) of mutually exclusive and exhaustive values. In an inference network, propositions typically can only take on the values "true" and "false." Another distinction is that influence diagrams represent uncertain relationships among propositions using the concept of *probabilistic dependency* while inference networks represent uncertain relationships using the concept of *belief update*.

The story of Mr. Holmes illustrates another difference between influence diagrams and inference networks. The top level of an influence diagram for Mr. Holmes' situation is shown in Figure 4.

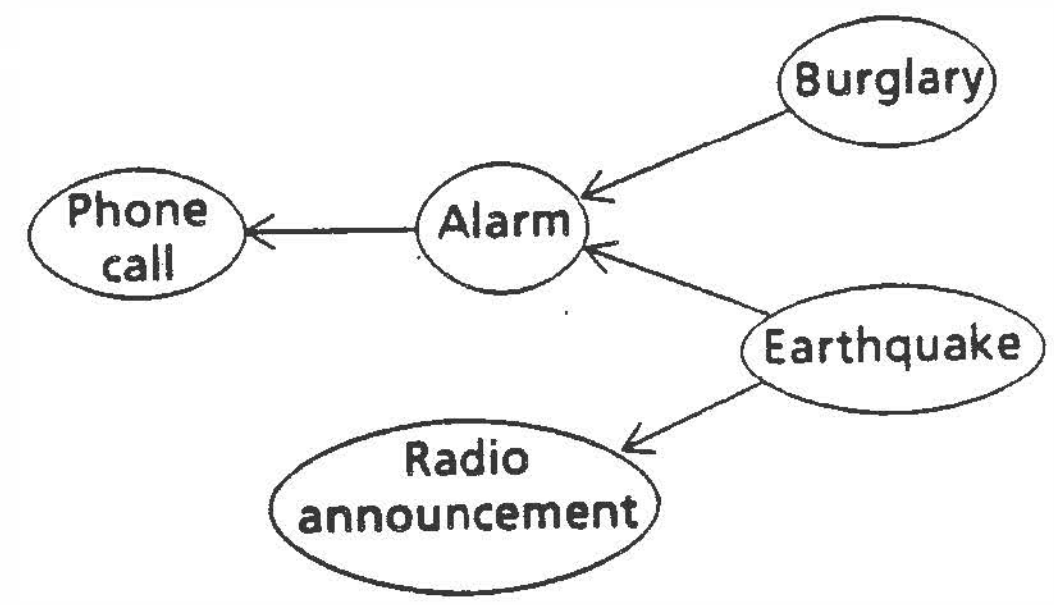

**Figure 4:** An influence diagram for Mr. Holmes' situation

Notice that many of the nodes in the graph are not directly connected by arcs. As mentioned earlier, the missing arcs are interpreted as statements of conditional independence. For example, the lack of a direct arc between "Burglary" and "Phone call" indicates "Burglary" influences "Phone call" only through its influence on "Alarm." In other words, "Burglary" and "Phone call" are conditionally independent given "Alarm." This would not be true if, for example, Mr. Holmes believed his neighbor might be the thief. We see from this example that influence diagrams provide a flexible means by which experts can assert assumptions of conditional independence that coincide with their beliefs. That is, assumptions of conditional independence are *imposed by the expert*. In contrast, assumptions of independence are *imposed by the methodology* in semantically modular inference networks.

Due to differences between inference networks and influence diagrams, problems that *cannot* be represented in the former approach can be represented in the latter. For example, the three urn problem could not be represented using an inference network because there were three mutually exclusive and exhaustive hypotheses. However, representing more than two mutually exclusive and exhaustive hypotheses is quite straightforward in influence diagrams. The problem of Mr. Holmes could not be represented using an inference network because of strong dependencies among "Alarm," "Burglary," and "Earthquake." In an influence diagram, however, these dependencies are naturally represented by the two arcs entering the "Alarm" node. In particular, since there are arcs from both the burglary and earthquake propositions to the alarm proposition, the second level of the influence diagram will contain the probability distribution for "Alarm" as a function all the possible values of both of these propositions. That is, the following probabilities will be included in the lower level of the influence diagram:

$p(Alarm | Burglary \wedge Earthquake)$

$p(\sim Alarm | Burglary \wedge Earthquake)$

$p(Alarm | Burglary \wedge \sim Earthquake)$

$p(\sim Alarm | Burglary \wedge \sim Earthquake)$

$p(Alarm | \sim Burglary \wedge Earthquake)$

$p(\sim Alarm | \sim Burglary \wedge Earthquake)$

p(Alarm|~Burglary∧~Earthquake)

p(~Alarm|~Burglary∧~Earthquake)

The interaction between the burglary and earthquake hypotheses is completely captured by this probability distribution.[6] In general, it can be shown if an inference problem can be solved in a decision-theoretic framework then it can be represented with an *influence diagram* [13]. As we have seen, this cannot be said for inference networks.

Now we are ready to consider a weaker notion of semantic modularity associated with the influence diagram representation. Imagine that a proposition is added to an influence diagram. When this occurs, the expert must first reassess the dependency structure of the diagram. For example, the new node may be influenced by other nodes, may itself influence other nodes, or may introduce conditional independencies or conditional dependencies among nodes currently in the network. Then, *the expert must reassess the probability distribution for each node which had its incoming arcs modified.* However, given the definition of an arc in an influence diagram, there is no need to modify the probability distributions for the nodes in the network whose incoming arcs were not modified. Similarly, if a proposition is deleted from an influence diagram, the expert must first reassess dependencies in the network and then modify only the probability distributions for those nodes which had their incoming arcs modified. The ability to add and delete propositions from an influence diagram without the need to reassign distributions for all nodes in the diagram is the weaker form of semantic modularity we have sought.

To illustrate the concept of weak semantic modularity, consider the following modification to Mr. Holmes' dilemma:

> Shortly after hearing the radio announcement, Mr. Holmes realizes that it is April first. He then recalls the April fools prank he perpetrated on his neighbor the previous year and reconsiders the nature of the phone call.

Given this new information, an "April fools" node should be added to the influence diagram as well as a conditioning arc from the new node to "Phone call." Furthermore, it appears that "April fools" directly influences only "Phone call" and that no other arcs need be added. Therefore, given the weaker form of semantic modularity we have outlined, only the probability distribution for "Phone call" need be reassessed; all other distributions remain intact. Note that the syntax of influence diagrams exactly parallels the new notion of semantic modularity we have defined. Thus, the influence diagram methodology is a framework in which the notions of semantic and syntactic modularity can be united in the context of plausible reasoning.

## Summary

In this paper, we have scrutinized the concept of modularity in detail. We have argued that this notion consists of two distinct concepts which we have called syntactic modularity and semantic modularity. We have argued that when reasoning under certainty, it is reasonable to regard the rule-based approach as semantically modular but when reasoning under uncertainty, semantic modularity rarely holds. To illustrate this point, we have examined the concept of semantic modularity in the context of the MYCIN certainty factor model and demonstrated that the assumption of semantic modularity entails both a strong conditional independence assumption and a strong restriction on the topology of the network. We have argued that such restrictions can rarely be met in practical applications and that consequently, semantic modularity does not hold in such situations. In addition, we have examined the implications of the semantic modularity assumption on knowledge engineering. Finally, we have described influence diagrams as a representation approach which accommodates a weaker form of semantic modularity appropriate for plausible reasoning.

## Acknowledgements

We wish to thank Judea Pearl and Peter Hart for several insightful discussions concerning divergence. We also thank Curt Langlotz, Ted Shortliffe, and Larry Fagan for their useful comments. Support for this work was provided by the Josiah Macy, Jr. Foundation, the Henry J. Kaiser Family Foundation, and the Ford Aerospace Corporation. Computing facilities were provided by the SUMEX-AIM resource under NIH grant RR-00785.

## Notes

[1]In another paper in this proceedings [3], the concept of a measure of change in belief or *belief update* is discussed in general terms.

[2]Technically, this assertion only holds for evidence e such that E does not lie on a directed path from e to H in the inference network. The certainty factor model explicitly handles the case where E lies on a directed path from e to H (see [9]).

[3]For simplicity, we will assume that prior information is included in $\xi_H$ and that pieces of evidence are logically distinct.

[4]Consistent propagation is possible if no other rules bear on $H_1$ or $H_2$.

[5]We note that the *Bayesian networks* of Pearl [2] and the *causal networks* of Cooper [1] are closely related to influence diagrams. In fact, all three approaches suggest the same form of semantic modularity.

[6]This example reveals a potential disadvantage of the influence diagram approach. In general, the number of probability assessments required for a single node is exponential in the number links converging on the node. However, Kim and Pearl [11] have developed a method whereby the probability distribution for a node can be calculated from "lower-order" probabilities in many situations. For example, their method can be used to calculate p(~Alarm|Burglary∧Earthquake) from p(~Alarm|Burglary) and p(~Alarm|Earthquake).